\documentclass[10pt, conference, compsocconf]{IEEEtran}
\let\labelindent\relax
\usepackage{cite}
\usepackage{amsmath,amssymb,amsfonts}
\usepackage{algorithmic}
\usepackage{graphicx}
\usepackage{textcomp}
\usepackage{xcolor}
\usepackage{times}
\usepackage{epsfig}
\usepackage{threeparttable}
\usepackage{graphicx}
\usepackage{amsmath}
\usepackage{textcomp}
\usepackage{hyperref}
\usepackage{import}
\usepackage{graphics}
\usepackage{booktabs}
\usepackage{caption}
\usepackage{subcaption}
\usepackage{enumitem}
\usepackage{tabularx}
\usepackage{multirow}
\usepackage{booktabs}
\usepackage{stfloats}
\usepackage{amssymb}
\usepackage{latexsym}
\usepackage{breqn}
\usepackage[utf8]{inputenc}
\makeatletter
\def\blfootnote{\xdef\@thefnmark{}\@footnotetext}
\makeatother

\newlength{\bibitemsep}
\newlength{\bibparskip}
\let\oldthebibliography\thebibliography
\renewcommand\thebibliography[1]{%
  \oldthebibliography{#1}%
  \setlength{\parskip}{\bibitemsep}%
  \setlength{\itemsep}{\bibparskip}%
}
\ifCLASSINFOpdf
\else
\fi
\begin{document}
\newcommand{\AFBsays}[1]{\textcolor{red}{[AFB says: #1]}}
\newcommand{\RGsays}[1]{\textcolor{blue}{[RG says: #1]}}
\newcommand{\MRsays}[1]{\textcolor{pink}{[MR says: #1]}}

%
\title{Selective Style Transfer for Text}


\author{\IEEEauthorblockN{Raul Gomez$^{a,b,*}$, Ali Furkan Biten$^{b,*}$, Lluis Gomez$^{b}$, Jaume Gibert$^{a}$, Marçal Rusiñol$^{b}$ and Dimosthenis Karatzas$^{b}$}
\IEEEauthorblockA{$^{a}$Eurecat, Centre Tecnol\`ogic de Catalunya, Unitat de Tecnologies Audiovisuals, Barcelona, Spain\\
$^{b}$Computer Vision Center, Universitat Aut\`onoma de Barcelona, Barcelona, Spain\\
\{ragomez,abiten,lgomez,marcal,dimos\}@cvc.uab.es, jaume.gibert@eurecat.org}
}

%

%


\maketitle

\blfootnote{$^{*}$Both authors contributed equally to this work.}

\begin{abstract}
This paper explores the possibilities of image style transfer applied to text maintaining the original transcriptions. Results on different text domains (scene text, machine printed text and handwritten text) and cross-modal results demonstrate that this is feasible, and open different research lines.
Furthermore, two architectures for selective style transfer, which means transferring style to only desired image pixels, are proposed.
Finally, scene text selective style transfer is evaluated as a data augmentation technique to expand scene text detection datasets, resulting in a boost of text detectors performance. Our implementation of the described models is publicly available\footnote{https://github.com/furkanbiten/SelectiveTextStyleTransfer}.
\end{abstract}


%
\IEEEpeerreviewmaketitle

\section{Introduction}
Style transfer is the task of combining the \textit{style} of one image with the \textit{content} of another image. Although the content of an image can be defined by the objects and the general scenery, the style of an image is not well defined. The style can be understood as the brush stroke of a painting, the color distribution, certain dominant forms and shapes or even a combination of all the above\cite{Gatys2015}.
Previous style transfer works have focused on transferring paintings styles, where the features to be transferred encode the brush strokes, the cubist patterns or the color palette achieving fascinating results\cite{Gatys2015,Johnson,Dumoulin}. 

However, text characters are very particular objects for which the common understanding of content and style cannot be adopted. Instead we define the style of the text as the shape, color and background of the characters and the content as the transcription of the text. In this work, we devise two architectures that learn the features that encode a certain text style and are able to transfer them to other text instances while preserving their content. We thus recast the ideas of style transfer, previously applied mostly to paintings, to the text domain.

Specifically, our method is able to automatically change the style of text regions in natural scene images, generating realistic images with the same textual content but with different text styles. In machine printed text images, we are able to train models that stimulate a change of the text font. In handwritten text, we are able to transfer the writing style of a particular writer to another. 

Possible applications of text style transfer include text stylization for augmented reality systems, or text font normalization in order to recast the visual appearance of any text instance to a canonical style in order to ease the different steps of a text recognition pipeline.


\begin{figure}[ht]
  \includegraphics[width=\linewidth]{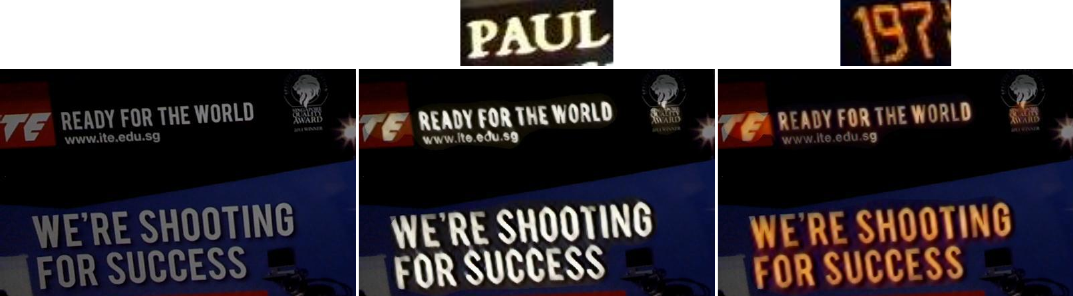}
     \caption{Original ICDAR 2015 images are shown on the left, and the resulting augmented images on their right.}
  \label{fig:icdar_style_transfer}
\end{figure}
In particular, we will demonstrate that such an approach is useful as a data augmentation technique (see Figure \ref{fig:icdar_style_transfer}), in order to deal with the problem of annotated data scarcity. 
One can transfer the styles found in a particular dataset to much bigger available annotated datasets, in order to train better text detectors or recognizers.

The contributions of this work are as follows:
\begin{itemize}[noitemsep,topsep=0pt]
   \item We demonstrate that existing style transfer pipeline can be used for text style transfer.
   \item We extend the prior works to selective style transfer for text, devising two architectures: two-stage and end-to-end, that allow to stylize only the text regions in an image.
    \item We use text style transfer as a data augmentation technique for the scene text detection task, and provide results that show a boost in the detection performance.
\end{itemize}
The rest of the paper is organized as follows: in the related work section, we overview the style transfer and the text stylizing literature. In the methodology section, we first define the style transfer task and then present the proposed two-stage and end-to-end architectures for selective text style transfer. In the results section, we explain the models we have trained in different text domains, show qualitative results and discuss their performance. Finally, in the data augmentation section, we show how text style transfer can be used to boost a text detector performance.
\section{Related Work}
Gatys \textit{et al.} \cite{Gatys2015} propose a method using a pretrained VGG network \cite{Simonyan15} for style transfer by extrapolating a randomly generated image to a stylized image. The stylized image is obtained by computing the backward propagation on the resulting pixel values which is computationally demanding. 
Johnson \textit{et al.} \cite{Johnson} recast the problem as an image transformation task, where a single, fixed learnt painting style is applied to an arbitrary image. A CNN is trained to alter a corpus of content images to match the style of a painting, eventually allowing to stylize images in real time. Simultaneously, Ulyanov \textit{at al.} \cite{Ulyanov2016} introduced the idea of Instance Normalization which is a modified version of Batch Normalization to have computationally less demanding models. A drawback of these works\cite{Johnson, Ulyanov2016} is that an independent model has to be trained for each source style. Dumolin \textit{et al.} \cite{Dumoulin} overcomes it by proposing a single CNN that can learn to transfer different source styles (up to 32 in their experiments), allowing to generate images with combined styles.


An appealing direction in the style transfer literature is to apply the style according to semantic segmentation. Li \textit{et al.}\cite{li2016combining} use a Markov Random Field over given semantic maps to decide which patch of the image to stylize. Luan \textit{et al.}\cite{luan2017deep} extend this idea to doodles with manual annotation. A final extension comes from Zhao \textit{et al.}\cite{zhao2017automatic}, in which they use two models: one for generating soft masks and the other for style transfer. All these works, however, apply the style transfer to the whole image. Conceptually, the closest to our work is that of Gatys \textit{et al.}\cite{gatys2017controlling} where they propose a method to control perceptual factors such as color, luminance and spatial location and scale. In particular, spatial control refers to applying the style to only masked areas which can be sky areas, balls, houses, trees, etc. The proposed method is \textit{guided Gram matrices} in which they use masking for the calculation of Gram matrices over CNN features at specific image regions. 

Although style transfer has not been applied to text, other works have targeted the task of changing text style with different approaches.
Liu \textit{et al.} \cite{Liu2018} proposed a pipeline to transform scene text into machine printed text within a scene text recognition model. 
Abe \textit{et al.} \cite{Abe2017} proposed a Generative Adversarial Network model to create new machine printed text fonts. Aksan \textit{et al.} \cite{AksanReb} propose a generative model to disentangle content and style of handwritten text represented as temporally ordered strokes, and apply it to handwriting synthesis and style transfer.
More related to our work, Ankan \textit{et al.} \cite{KumarBhunia2018} focus on font to font translation in images of printed documents using a GAN architecture. Also, Azadi \textit{et al.} \cite{Azadi2018} propose a conditional GAN to style machine printed text to more complex scene text fonts, learning each character style independently.

Our work, to the best of our knowledge, is the first one exploring the performance of style transfer models on the text style transfer task as well as having an end-to-end model that can perform spatial control without any external information. 


\section{Methodology}
We will first explain the details of the model on which we base our implementation. Afterwards, we will explain the details of our architectures, a two-stage and an end-to-end model that apply style transformations to only text areas in the image. 
\subsection{Style Transfer} 

We use the model proposed by Dumoulin \textit{et al.}\cite{Dumoulin} as our baseline. The style transfer task is usually defined as finding an image $p$ which is produced by an encoder-decoder image transformation network, whose content is similar to a source content image $c$ but whose style is similar to a source style image $s$. A key point of style transfer is the definition of both content and style. In \cite{Dumoulin}, two images are considered to have a similar content if their high-level features extracted by a trained classifier are close in Euclidean distance. On the other hand, two images are similar in style if their gram matrices of low-level features as extracted by a trained classifier are close under the Frobenius norm.
More formally, 
let $\phi$ be the transformer network and $\gamma^l$ the output of the $l^{th}$ layer of a CNN pretrained on ImageNet\cite{Deng}. In our case $\gamma$ is the VGG-16\cite{Simonyan15}. The training process is as follows: we initially forward the content image $c$ through the transformer network $\phi$ to obtain the stylized image $p$. All images $c$, $s$, $p$ are then forwarded through $\gamma$, and features for them are extracted: correspondingly, $F_c$ for the content image from the $m^{th}$ layer, $F_s$ for the style image from $n^{th}$ layer and, $F_{pm}$, $F_{pn}$ for the stylized image from both $m^{th}, n^{th}$ layers where $m\geq n$. The content loss $L_c$ is defined as the mean squared error between $F_c$ and $F_p$. The style loss $L_s$ is computed as the mean squared error between corresponding Gram matrices $G_s$ and $G_p$ of the features $F_s$ and $F_p$. The final loss $L_{total}$ that directs the model training is a weighted average of the content and the style losses. In summary:
\begin{equation}
\begin{split}
    &
p = \phi(c),\\&
F_c = \gamma^m(c), F_s = \gamma^n(s),\\&
F_{pm} = \gamma^{m}(p), F_{pn} = \gamma^{n}(p),\\& 
L_c = \sum_k{(F_{ck} - F_{pmk})^2}, \\&
G_s = F_s \cdot F_s^T, G_p = F_{pn} \cdot F_{pn}^T \\&
L_s = \sum_k{(G_{sk} - G_{pnk})^2}, \\&
L_{total} = \lambda_1  L_c + \lambda_2  L_s.
    \end{split}
\end{equation}


\subsection{Selective Style Transfer for Text} 

Selective style transfer refers to automatically detecting the relevant areas in the image (in our case, areas where text is present) and restricting the application of style to the detected areas only, leaving the rest of the image unchanged.
Accordingly, we design and describe two models that can perform selective style transfer for text.


\subsubsection{Two-stage architecture}

\begin{figure}[ht]
  \includegraphics[width=\linewidth]{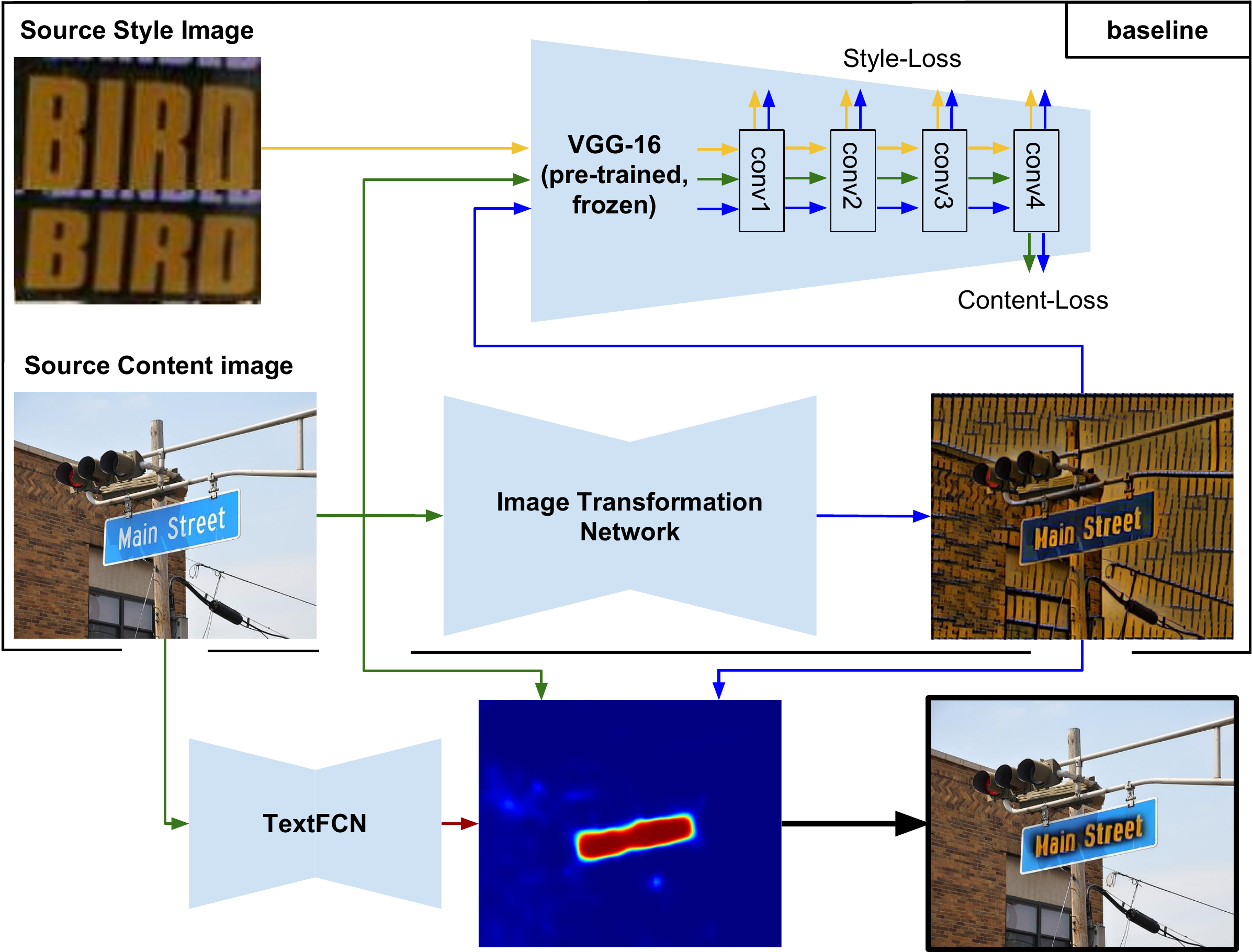}
     \caption{Two-stage Selective Text Style Transfer pipeline.}
  \label{fig:pipeline}
\end{figure}

\begin{figure}[ht]
  \includegraphics[width=\linewidth]{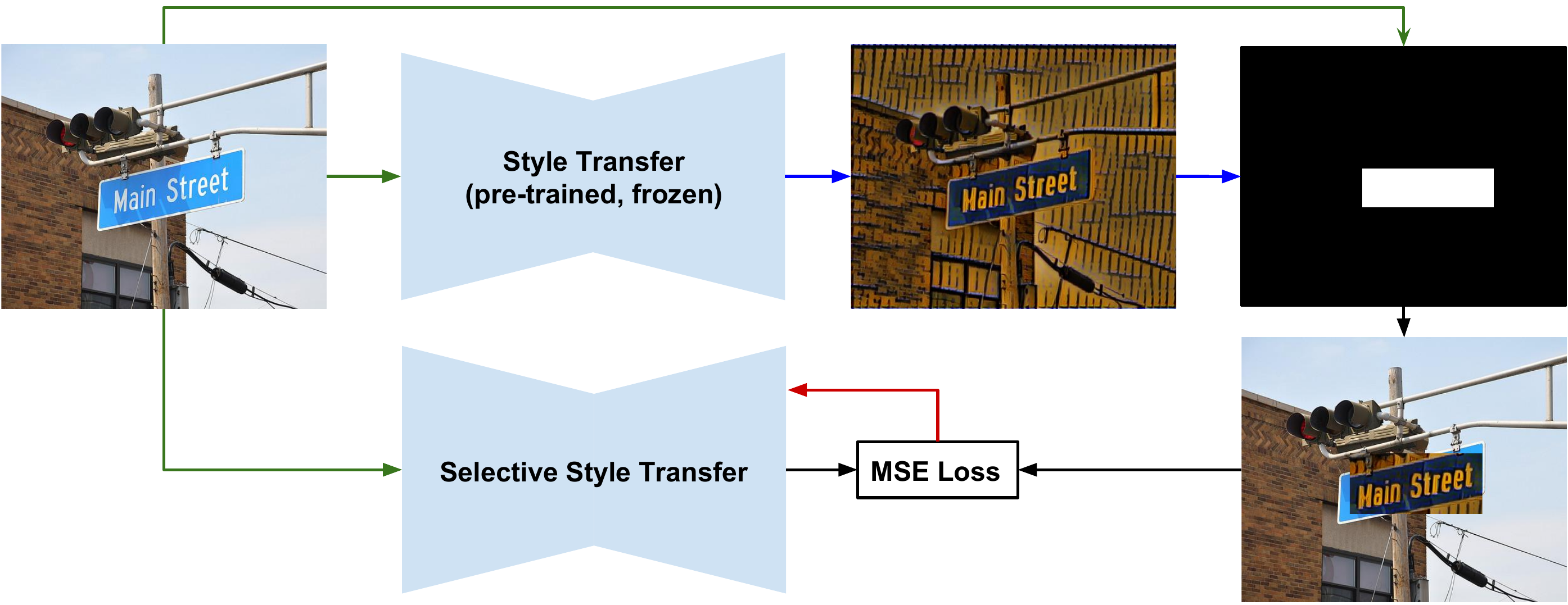}
     \caption{End-to-end Selective Text Style Transfer pipeline.}
  \label{fig:end2end_training}
\end{figure}

To stylize only textual areas of the images, we exploit TextFCN \footnote{https://github.com/gombru/TextFCN} \cite{Bazazian2017, Bazazian2017a} which is a text detector that infers the probability of each pixel belonging to a textual area. To transfer a text style to an input image, we first stylize the whole image. Then, we compute the pixel-level heatmap of the original image with TextFCN. To generate the final image where the style is transferred only to textual areas, we do a blending of the original image and the stylized image weighted with the TextFCN heatmap (see \autoref{fig:pipeline}). 
This procedure allows to obtain realistic images, ensuring that non-textual areas are kept unchanged. 

More formally, 
given a stylized image $p$ and a pre-trained TextFCN $\delta$, we process the content image $c$ with $\delta$ to obtain the per-pixel text probability map $P_t$. 
Then we get only textual areas of the stylized image $p$ by taking its Hadamard product with $P_t$, and do the same with $c$ and $1 - P_t$ to get the content of non-textual areas. We sum up the results to get the final image $p_{text}$:

\begin{equation}
    \begin{split}
&
P_t = \delta(c), \\ &
p_{text} = P_t \odot p + (1 - P_t) \odot c
    \end{split}
\end{equation}

\subsubsection{End-to-end Architecture}
In the style transfer literature, pre-computed masks of image regions have been extensively used to apply different styles in different regions\cite{gatys2017controlling, luan2017deep,li2016combining,zhao2017automatic}. However, masks are used 
to constrain the image transformation network output
as we do with the two-stage architecture, or for gradient masking, and none of the existing works is capable of learning the masks along with the style information. To this end, combined with the purpose of reducing the computational complexity, we create a novel end-to-end architecture that is capable of performing selective style transfer on text without needing any text detector. Our model is inspired by the distillation strategy from \cite{hinton2015distilling}. 
The basic idea of distillation is to pass the learnt information of various networks, which are able to solve different tasks, into a single model. In our case, we combine the image style transformation network with the text detector. We take the pretrained image style transformation network and the ground truth annotations for the text to train a randomly initialized image transformation network with mean squared error loss (see Figure \ref{fig:end2end_training}). 

More formally, let $\phi$ be the pretrained image style transformation network, $M$ the masks for the text regions where $M_{ij} \in \{0,1\}$ and, $\eta$ the same network as $\phi$ but randomly initialized. 
We first obtain the stylized image $p$ forwarding the content image $c$ though $\phi$. We then obtain $\hat p_{text}$ as the output of $\eta$ after feeding it with the content image $c$. To get the ground truth for content and for style, $\theta_c$ and $\theta_s$ respectively, we apply the Hadamard product:   
\begin{equation}
    \begin{split}
    & p = \phi(c), \\ 
    & \hat p_{text} = \eta(c), \\ 
    & \theta_{s} =  p \odot M, \\ 
    & \theta_{c} = c \odot (1 - M). \\ 
    \end{split}
\end{equation}

We use mean squared error as our loss to train the selective style transfer net $\eta$. There are two key points in our loss calculation. First of all, we need to apply the mask $M$ for text regions, and $1-M$ for content regions to $\hat p_{text}$ to make sure our model learns to differentiate between text and background. Secondly, since text regions in the image are significantly smaller compared to background, we weight loss contributions of text and background pixels with two parameters $\lambda_1$, $\lambda_2$, with $\lambda_1 > \lambda_2$. The loss is defined as:
\begin{multline}
L  = \lambda_1  \sum(\hat p_{text} \odot M - \theta_s)^2 + \\ 
  \lambda_2  \sum(\hat p_{text} \odot (1 - M) - \theta_c)^2.
\end{multline}

\section{Experiments}

\begin{figure}[ht]
  \includegraphics[width=\linewidth]{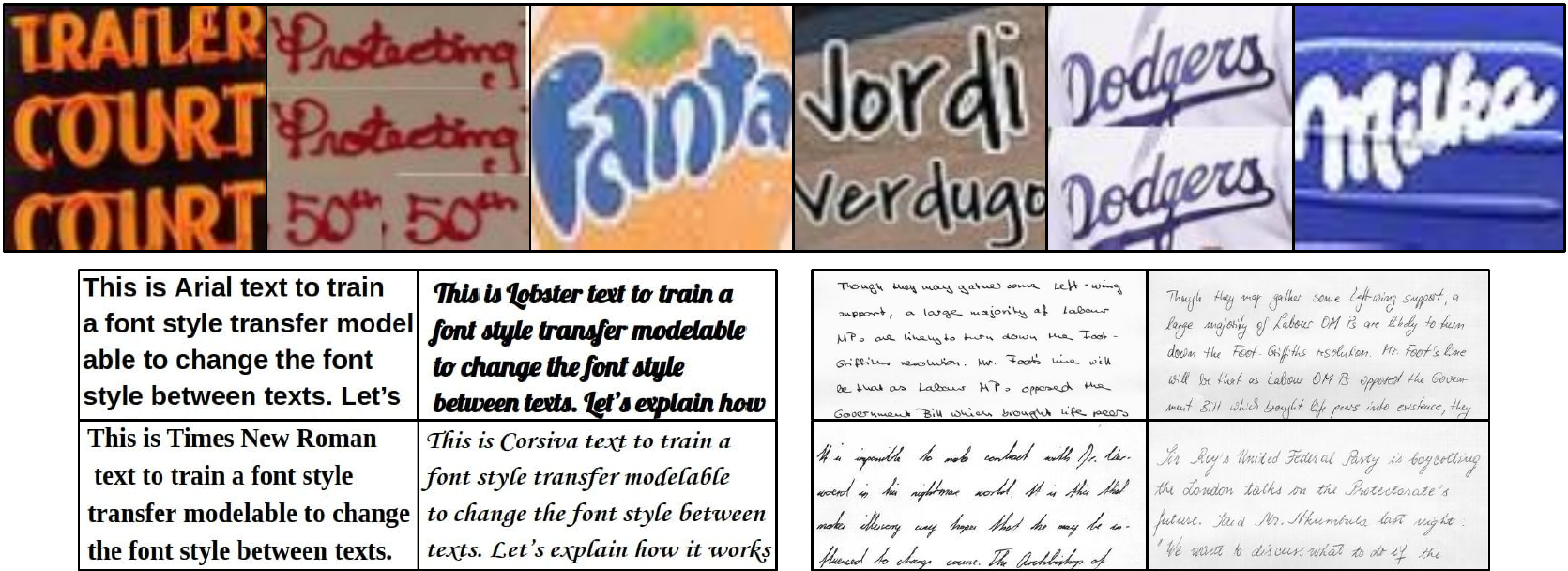}
     \caption{Some of the source styles for the scene text model, the machine text model and the handwritten text model.}
  \label{fig:src_styles}
\end{figure}

To explore the capabilities of text style transfer in various text domains, we train 3 models, namely, a model to transfer scene text styles, a model to transfer machine printed text fonts styles, and a model to transfer handwritten styles. Figure \ref{fig:src_styles} shows some of the source styles used. In this section, we explain the implementation and training details for each model, and give qualitative results.

\subsection{Scene Text}

We trained our baseline style transfer model for two-stage architecture with 34 scene text styles from COCO-Text dataset \cite{Veit16} using cropped word images as source styles and ImageNet \cite{Deng} as the training dataset. Then, we trained our end-to-end model for selective text style transfer using the COCO-Text \cite{Veit16} dataset (only the images containing legible text). To train the end-to-end model in the distillation fashion, we set  parameters $\lambda_1 = 100, \lambda_2 = 1$ and trained for $77$ epochs. After that, we got images where the style was correctly transferred to text, but were quite blurry and dark. Then we trained for $50$ epochs with equal weight for textual and not textual areas to balance the style and content loss.

\begin{figure*}
\centering
  \includegraphics[width=\linewidth]{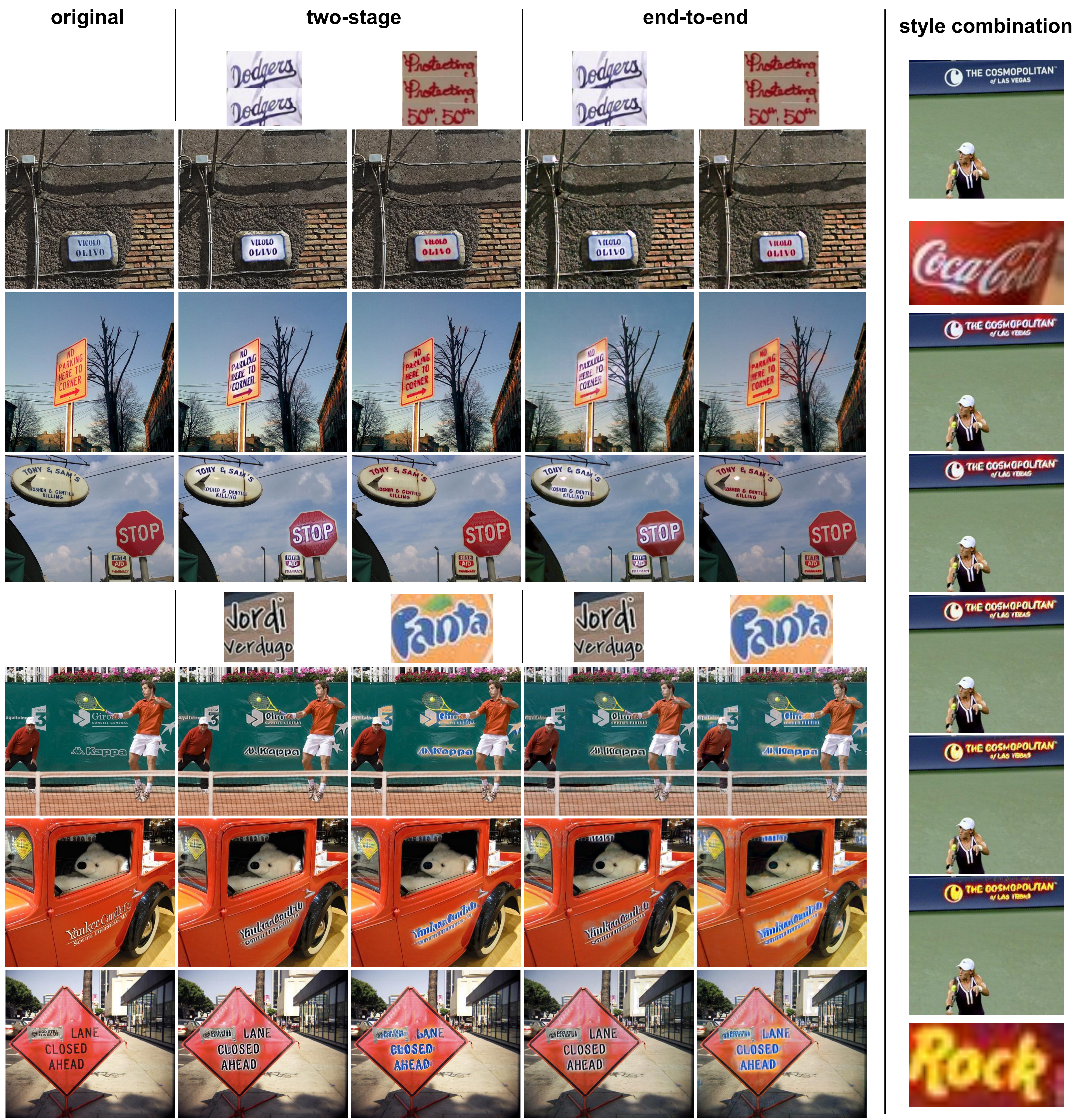}
     \caption{Applying different styles to various scene text images using the two-stage and end-to-end architectures with the cropped word styles (left).
     Extrapolating from ``Coco-Cola'' to ``Rock'' style on COCO-Text images (Right). (Best viewed in color.)}
  \label{fig:results_scene_text_all}
\end{figure*}

Figure \ref{fig:results_scene_text_all} shows results of the scene text model applied to COCO-Text scene images, using both the two-stage architecture and the end-to-end architecture to get the selective text style transfer results. The performance is appealing, transferring the source styles with high fidelity in both character shapes and colors to a wide diversity of scene texts. The text content is preserved quite well in most images, and only in some cases where the task is very complex due to the original text size or tangled style the result is illegible. Both the weight blending with TextFCN and the end-to-end architecture allow getting realistic images that are very useful as data augmentation (see Section \ref{sec:data_augmentation}). This realistic results have also a huge potential in artistic applications, such as graphic design. Graphic designers could quickly visualize how a given text style fits in a existing scene text. Moreover, the model can perform style transfer averaging several styles, which allows to create new styles.




\subsection{Machine Printed Text}
We train our baseline model with 8 machine printed source text fonts: Arial, Times, Lubster, Corsiva, Caveat, Pacifico, Consolas and Syncopate and Imagenet \cite{Deng} training images. Figure \ref{fig:results_ht_mt} shows results of the model applied to machine text. It transfers successfully the main features of the source font style, such as line width, text orientation, and main font character style. However, it fails transferring the specific styles of some characters, and the final output is influenced by the initial image. 


\subsection{Handwritten text}
\begin{figure*}
\centering
  \includegraphics[width=\linewidth]{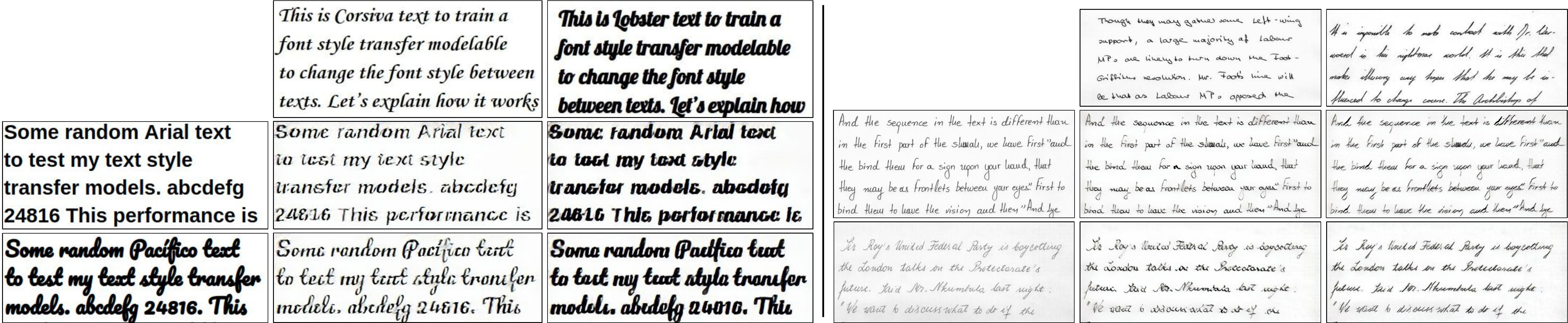}
     \caption{Results of the machine printed text model (left) and handwritten model (right). In each text domain, the styles of the images on top are transferred to the images on the left.}
  \label{fig:results_ht_mt}
\end{figure*}

The baseline model is trained with 8 styles from different writers, using images from the IAM dataset \cite{Marti2002} as source styles and the ImageNet \cite{Deng} dataset as training images. Figure \ref{fig:results_ht_mt} shows results of the model applied to IAM dataset images. The model transfers correctly the main features of the text, as the tight characters and the thick stroke of the style in the first column, and the elongated and italic style of the writer in the second column. However, it fails on transferring more fine-grained characteristics of the source writer style, and some words of the resulting text are blurry.


\begin{figure}[ht]
  \includegraphics[width=\linewidth]{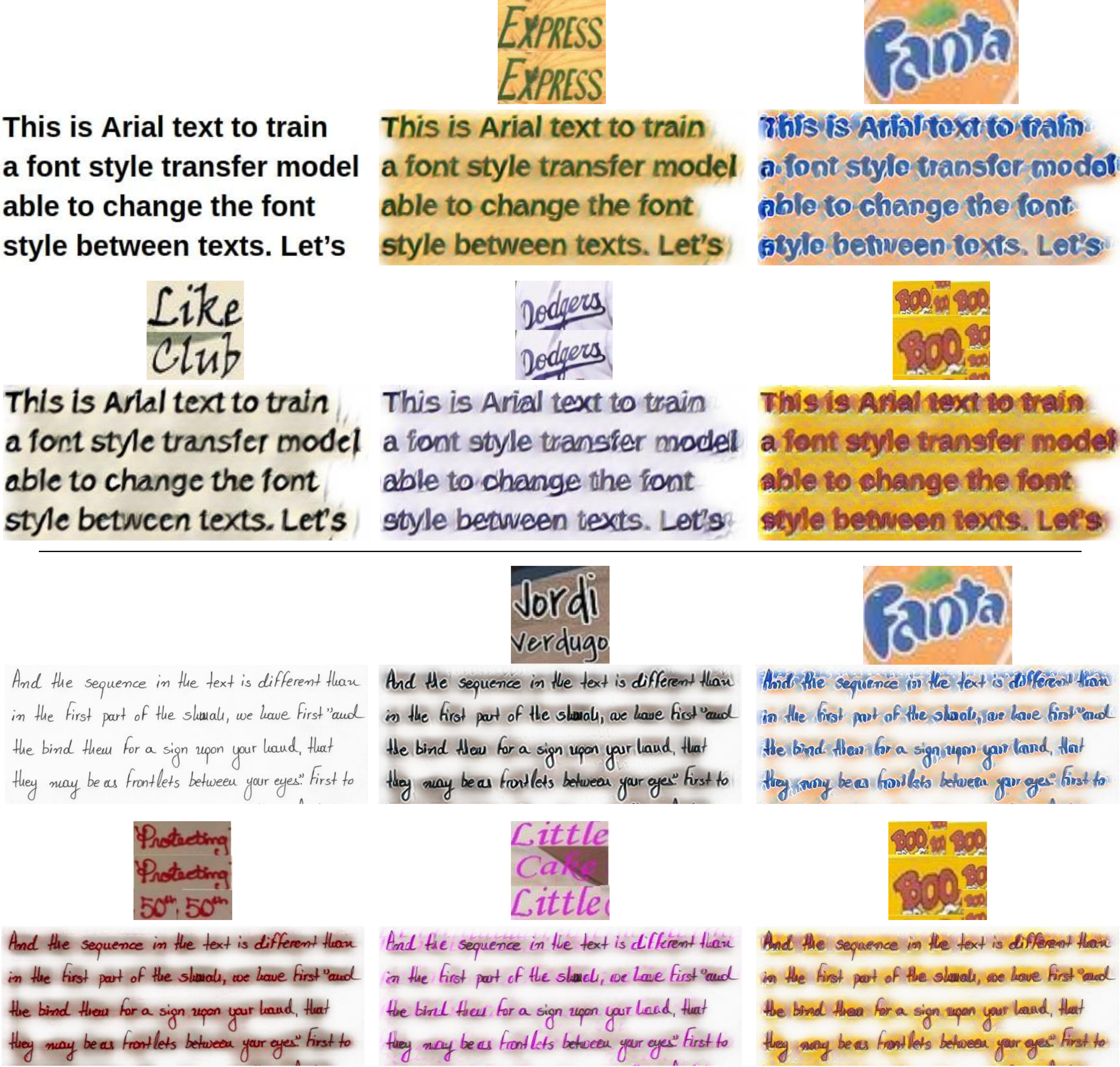}
     \caption{Results of the scene text model applied to machine printed text (top) and handwritten text (bottom) images.}
  \label{fig:machine_hw_2scene}
\end{figure}
\begin{figure}[ht]
  \includegraphics[width=\linewidth]{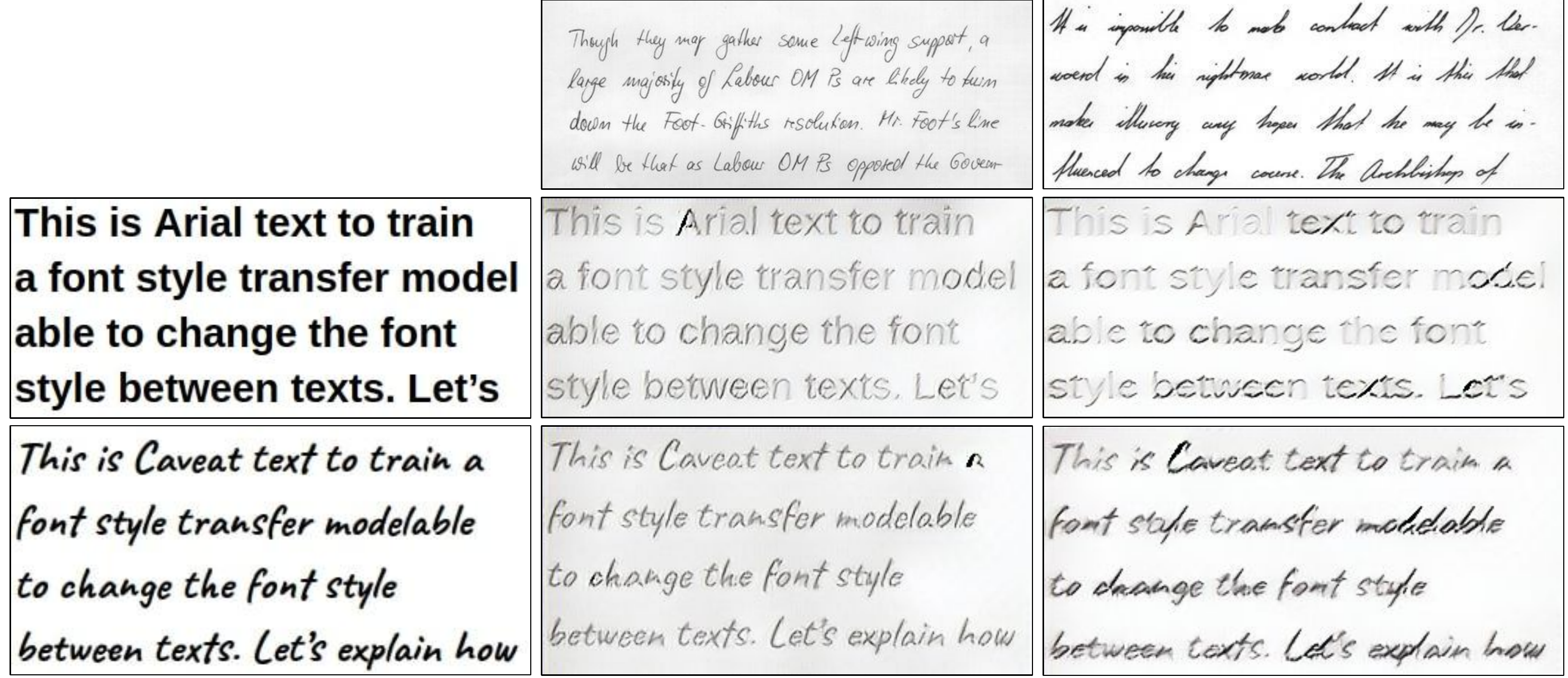}
     \caption{Handwritten text model transferring styles (top) to machine text images (left).}
  \label{fig:results_machine2handwritten}
\end{figure}
\begin{figure}[ht]
\centering
  \includegraphics[width=0.9399\linewidth]{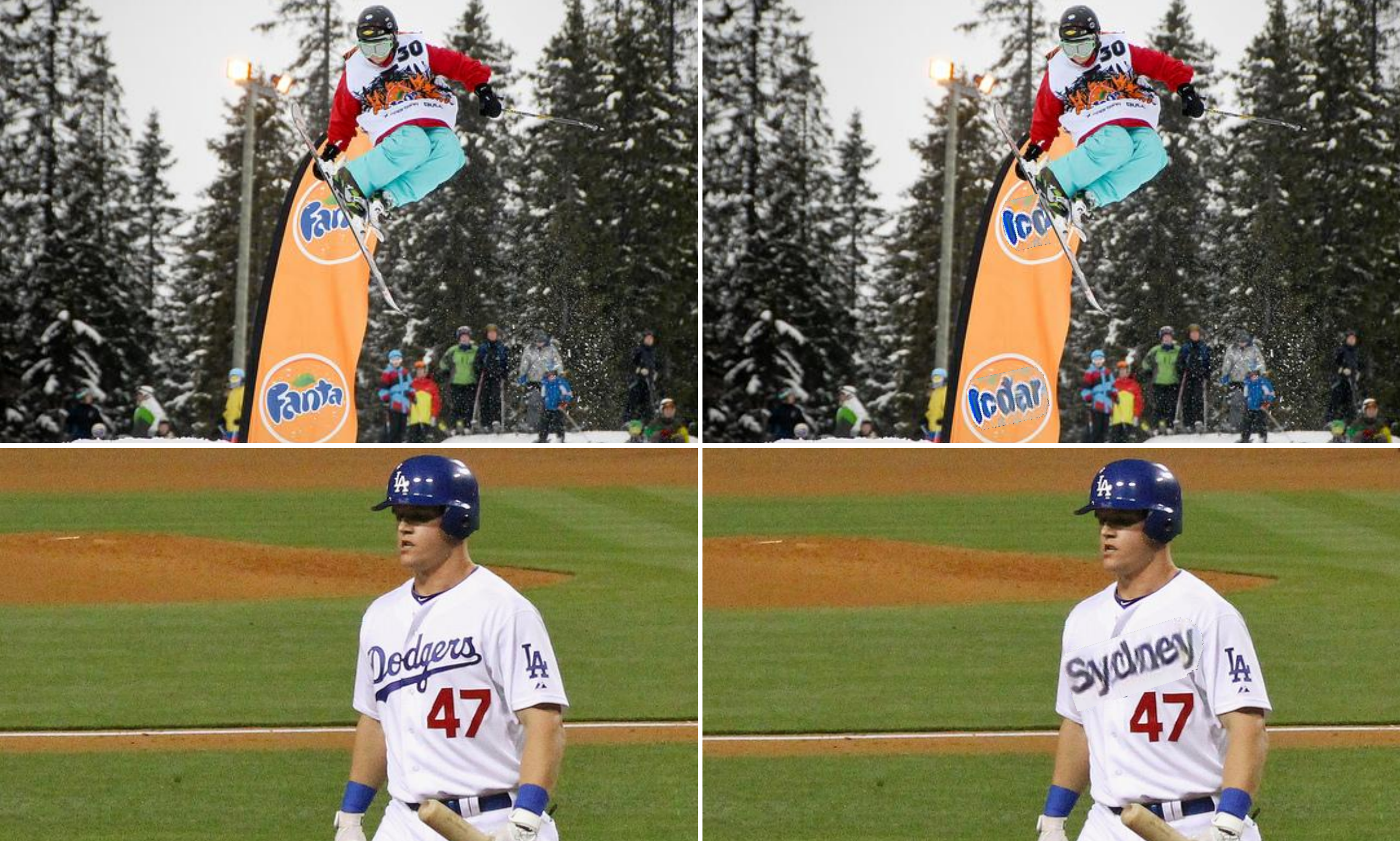}
     \caption{Arial text has been stylized with the original image scene text style (left) and manually inserted (right).}
  \label{fig:content_augmentation}
\end{figure}
\begin{figure}[ht]
  \includegraphics[width=\linewidth]{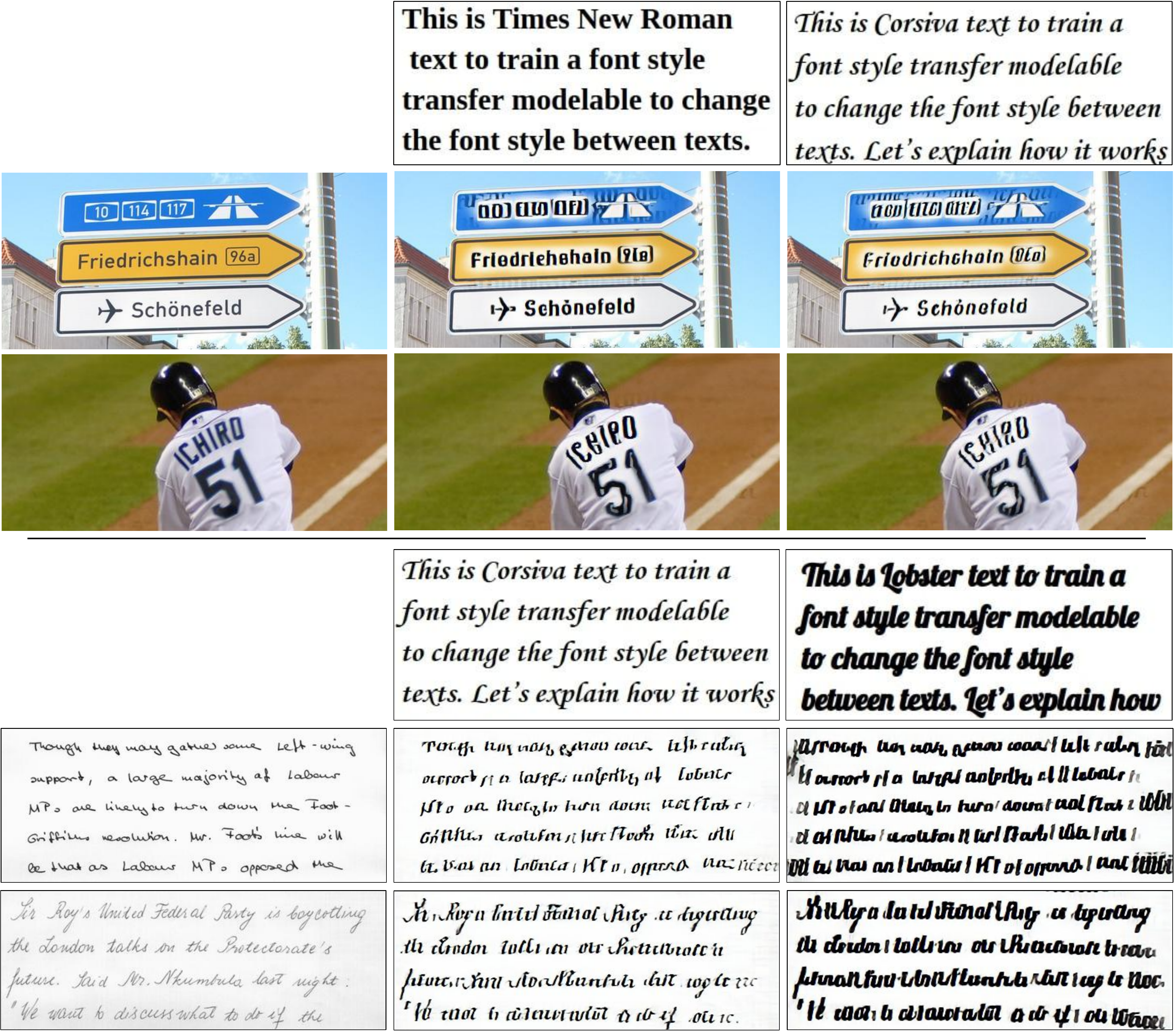}
     \caption{Results of transferring machine printed text styles to scene images (top) and handwritten images (bottom).}
  \label{fig:scene_hw_2machine}
\end{figure}

\subsection{Cross Domain}
In this section, we go one step further and test the capability of our text style transfer models to transfer style to images of other text domains.
\subsubsection{Machine Text to Scene Text}
Transferring scene text styles to machine printed text has a huge potential in augmented reality scenarios and as a data augmentation technique, generating synthetic images with a given text style but different text content.
Figure \ref{fig:machine_hw_2scene} shows results of styling machine printed text with the scene text model. The model successfully transfers scene text style to machine printed text with high fidelity.

Figure \ref{fig:content_augmentation} shows some content augmentation results, where Arial text has been stylized with the destination scene text font, and the stylized text has been inserted in the image manually.

\subsubsection{Handwritten to Scene Text}
Figure \ref{fig:machine_hw_2scene} shows results of styling handwritten text with the scene text model. 
This task is quite complex, since the scene text tends to be thick and detached while handwritten text is tangled formed by tangled thin strokes. However, the scene text style transfer model successfully transfers some style features to it keeping it legible, and results are a combination of the transferred scene text style and the original handwritten style.

\subsubsection{Scene Text to Machine Text}
A model capable to convert any scene text to machine printed text would be a nice tool to improve scene text understanding pipelines. The machine printed text model allows us to do so, as shown in Figure \ref{fig:scene_hw_2machine}. It correctly transfers the machine text source style if the scene text is simple, but it fails when scene text has a complex font, is too small or rotated. Note that the artifacts in those images are due to noisy high responses of the text detector.

\subsubsection{Handwritten to Machine Text}
Converting handwritten text to machine printed text could be very useful in a handwritten text understanding pipeline. 
Our machine text model transfers style features from machine fonts to handwritten text, but it breaks the content, resulting in illegible images, as shown in Figure \ref{fig:scene_hw_2machine}.

\subsubsection{Machine Text to Handwritten}
Converting machine text to handwritten text can be a great tool to generate synthetic data to train handwritten text understanding models. However, our handwritten text model fails transferring the font styles to machine text, as shown in Figure \ref{fig:results_machine2handwritten}. It only achieves to copy some general handwritten style features to some machine text fonts closer to handwritten styles, like Caveat. 



\section{Data Augmentation} \label{sec:data_augmentation}
In this section we include experiments that demonstrate the usefulness of the proposed selective scene text style transfer as a data augmentation tool to improve text detectors' performance. The reason for using style transfer as a data augmentation technique is twofold. According to the recent seminal work by Geirhos \textit{et. al.} \cite{geirhos2018imagenet}, CNNs are strongly biased towards recognising textures rather than shapes. To overcome the biases, they have made use of neural style transfer as an augmentation technique which resulted not only in overcoming the biases that a CNN has but also improving results on ImageNet classification\cite{Deng}. Thus, data augmentation using selective style transfer is expected to help overcome similar biases of the text recognizer. A second rationale is that some datasets do not offer the necessary amount of images to efficiently train large neural networks. Style transfer in such cases can artificially increase the size of the datasets by combining different styles and contents. This is especially true for ICDAR 2013 and ICDAR 2015. 

Moreover, the proposed data augmentation technique has a clear benefit compared with other methods to generate synthetic data \cite{Gupta2016,Zhan2018}. The generated images contain a text with a different visual style, but the text appears in the same place as in the original images, which makes the text position in the image realistic while preserving the content of the text, which makes the transcription match the scene semantics.

In the experiments, we use the consolidated and widely used EAST \cite{Zhou2017} text detector\footnote{\url{https://github.com/argman/EAST}}.
We consider the following datasets:
\begin{itemize}[noitemsep,topsep=0pt]
\item \textbf{ICDAR 2013} \cite{Karatzas2013}: the dataset contains $229$ training images and $229$ test images that capture focused text on sign boards, posters, etc.
\item \textbf{ICDAR 2015} \cite{Karatzas2015}: the dataset contains $1000$ training images and $500$ testing images  with incidental scene text, which means text that appears in the scene without the user focusing on it. 
\item \textbf{COCO-Text} \cite{Veit16}: the dataset contains $63k$ images from the COCO \cite{Lin2014} dataset with text regions annotated.
\end{itemize}

For these experiments, we trained an additional scene text style transfer model using $96$ different styles from the ICDAR 2015 training dataset, and used the two-stage architecture to perform selective text style transfer. 
We augment ICDAR 2013, ICDAR 2015 and COCO-Text datasets using the two-stage model with $1$ or $4$ random additional styles per image. 
Figure \ref{fig:icdar_style_transfer} shows some of the ICDAR 2015 resulting augmentations. We train the EAST text detector on the augmented and regular datasets. Stylizing COCO-Text with ICDAR 2015 training styles, allows to get a COCO-Text dataset closer to the target testing data, which is ICDAR 2015 testing set. To evaluate the trained models, we use the Robust Reading Competition framework (ICDAR 2015 Challenge 4: Incidental Scene Text Localization task 
and ICDAR 2013 Challenge 2: Focused Scene Text). 
Results in Table \ref{tab:results} show that text style transfer is a useful data augmentation technique, achieving an improvement in F-Score performance between $2$-$4$\% in all the setups just using $1$ or $4$ augmentations per image.


The boost in performance confirms that neural style transfer offers an advantageous and practical data augmentation technique that works both within the datasets as well as cross dataset, domain adaption scenarios.

\begin{table}[ht]
\centering
\begin{tabular}{cccc}
\toprule
\textbf{Training Dataset} & \textbf{Testing dataset} & \textbf{Augmentations} & \textbf{F} \\
\midrule
ICDAR 13 & ICDAR 13 & \multicolumn{1}{c}{-} & 70.97           \\ 
ICDAR 13 & ICDAR 13 & 1 styles per image & 74.55   \\ 
ICDAR 13 & ICDAR 13 & 4 styles per image & \textbf{75.29}   \\ \midrule
ICDAR 13+15 & ICDAR 15 & \multicolumn{1}{c}{-} & 80.83*           \\ 
ICDAR 15 & ICDAR 15  & \multicolumn{1}{c}{-} & 78.74            \\ 
ICDAR 15 & ICDAR 15   & 1 style per image      & 80.60            \\ 
ICDAR 15 & ICDAR 15   & 4 styles per image     & \textbf{81.83}   \\ \midrule
COCO-Text  & ICDAR 15   & \multicolumn{1}{c}{-} & 68.05            \\ 
COCO-Text  & ICDAR 15   & 1 style per image      & 69.66            \\ 
COCO-Text  & ICDAR 15   & 4 styles per image     & \textbf{70.71}   \\ \bottomrule
\end{tabular}
\caption{Results (F-score) of the EAST text detector with different training data, evaluated on ICDAR 2015 Challenge 4 and ICDAR 2013 Challenge 2. *Result reported on the original EAST github.}
\label{tab:results}
\end{table}


\section{Conclusions}
We have shown that a style transfer model is able to learn text styles as the characters shapes, line style, and colors, and to transfer it to an input text preserving the original characters.
We have explored the performance of text style transfer in 3 text modalities: scene text, machine printed text and handwritten text and in cross-modal scenarios, proving the usefulness of text style transfer as a data augmentation technique to train scene text detectors. 
Cross-modal experiments show the potential of this pipeline in virtual reality scenarios to style an arbitrary text with a given scene style, and the realism of the generated images suggest that the pipeline could be useful in artistic applications, such as graphic design.
We open the field for further research in different directions, such as data augmentation for scene text detection or recognition or handwritten writer identification. 

The proposed selective style transfer two-stage and end-to-end architectures allow to automatically get realistic images where only text has been styled. Furthermore, the end-to-end selective style transfer pipeline can be applied in other style transfer tasks besides text.
We provide PyTorch style transfer code, based on Google's TensorFlow Magenta implementation, including the end-to-end selective style transfer implementation, and text style transfer trained models for both frameworks.

\section*{Acknowledgments}
This work was supported by projects TIN2017-89779-P,
Marie-Curie (712949 TECNIOspring PLUS), aBSINTHE (Fundacion BBVA 2017), Doctorats Industrials (AGAUR), the CERCA Programme / Generalitat de Catalunya, NVIDIA Corporation and a UAB PhD scholarship.




%

\bibliographystyle{plain}
\bibliography{main}

\end{document}